\documentclass{article}
\UseRawInputEncoding
\usepackage{PRIMEarxiv}
\usepackage{amsmath}
\usepackage[utf8]{inputenc} 
\usepackage[T1]{fontenc}    
\usepackage{hyperref}       
\usepackage{url}            
\usepackage{booktabs}       
\usepackage{amsfonts}       
\usepackage{nicefrac}       
\usepackage{float} 
\usepackage{microtype}      
\usepackage{lipsum}
\usepackage{fancyhdr}       
\usepackage{graphicx}       
\graphicspath{{media/}}     
\usepackage{multirow}

\title{  Serialized Point Mamba: A Serialized Point Cloud Mamba Segmentation Model
}

\author{
  Tao Wang, Wei Wen,Jingzhi Zhai,Kang Xu,Haoming Luo \\
  Hunan University\\
  \texttt{\{wangtao,wwinfo\}@hnu.edu.cn} \\
}

\begin{document}
\maketitle

\begin{abstract}
Point cloud segmentation is crucial for robotic visual perception and environmental understanding, enabling applications such as robotic navigation and 3D reconstruction. However, handling the sparse and unordered nature of point cloud data presents challenges for efficient and accurate segmentation. Inspired by the Mamba model's success in natural language processing, we propose the Serialized Point Cloud Mamba Segmentation Model (Serialized Point Mamba), which leverages a state-space model to dynamically compress sequences, reduce memory usage, and enhance computational efficiency. Serialized Point Mamba integrates local-global modeling capabilities with linear complexity, achieving state-of-the-art performance on both indoor and outdoor datasets. This approach includes novel techniques such as staged point cloud sequence learning, grid pooling, and Conditional Positional Encoding, facilitating effective segmentation across diverse point cloud tasks. Our method  achieved 76.8 mIoU on Scannet and 70.3 mIoU on S3DIS. In Scannetv2 instance segmentation, it recorded 40.0 mAP. It also had the lowest latency and reasonable memory use, making it the SOTA among point semantic segmentation models based on mamba.

\end{abstract}


\section{Introduction}
\textbf{3D point cloud segmentation} is the process of classifying and grouping the points within a three-dimensional dataset captured by sensors such as LiDAR or depth cameras. By analyzing attributes such as the spatial coordinates, color, and intensity of each point, segmentation algorithms divide the point cloud data into regions or objects with distinct semantic meanings. These regions can represent various objects or scene elements, such as buildings, vehicles, pedestrians, and vegetation. 3D point cloud segmentation is a crucial technology in fields like robotic vision, environmental modeling, and autonomous driving, enhancing the system's ability to understand and interact with complex environments. 

Unlike image pixels or 3D voxels, much of the space in a 3D point cloud is empty, which means that point clouds are sparse data and lack specific order. Therefore, for certain transformations of 3D point clouds such as rotations and translations, the interpretation of the point cloud by 3D segmentation model should not change. Despite the discrete and sparse nature of the points, they are not isolated. Structures formed by neighboring points can represent meaningful subsets. Thus, a segmentation model needs to be capable of learning  the local features of the point cloud, and the relationship between local features and the overall context. On the other hand, when dealing with large-scale point clouds, efficiency and real-time performance often come at the cost of sacrificing precision and accuracy, which is typically challenging to reconcile. 

In recent years, the Transformer architecture has made significant advances in natural language, images, videos, and 3D data processing. As the core module of the Transformer, the self-attention mechanism can model complex data and handle long-sequence dependencies. However, the self-attention mechanism cannot model content beyond a finite window, and it requires explicit storage of the entire context (key-value cache). Consequently,  memory and computational consumption  grow quadratically with the input sequence length. Numerous studies have developed more efficient attention variants\cite{tay2022efficient}.  \textbf{ Flash attention  reduces computational complexity for long sequences, but  it may limit the model's ability to capture global context.  RWKV reduces memory and computational costs , but it increases system complexity.   } Another class of models, traditional RNNs, compress the context into a limited state, with computation and memory usage growing linearly with sequence length. However, they struggle to learn long-range dependencies.

State Space Models (SSMs)\cite{gu2023mamba}\cite{gu2021efficiently} have recently demonstrated their capability to simulate long-term dependencies in sequence data. Mamba\cite{gu2023mamba} has shown comparable or even superior performance to Transformers in several challenging NLP tasks. Inspired by Mamba's success in language modeling, we  leverage Mamba to design a generic and efficient backbone for point cloud segmentation. However, due to the unique structural properties of point cloud data, Mamba's sequence modeling capabilities cannot be directly applied to point clouds.

Addressing these issues, this paper investigates novel point cloud semantic segmentation algorithms. We propose the Serialized Point Cloud Mamba Segmentation Model (Serialized Point Mamba), employing a purely input-aware structured state space model to dynamically compress sequences. The model reduces memory usage and computation costs, and improves segmentation performance. To tackle the unordered nature of point clouds, we explore methods to restructure point cloud data for sequential modeling, avoiding the use of complex data structures and operations. We also utilize an enhanced positional encoding to reinforce the understanding of sequences. For large-scale point cloud data, a staged approach for local point cloud sequence learning is adopted, making full-process modeling with Mamba feasible. Various serialization techniques are employed to facilitate interaction among learned local features from different subsequences.

Our main contributions can be summarized as follows:
\begin{enumerate}
    \item We introduced the Serialized Point Cloud Mamba Segmentation model (Serialized Point Mamba). The model combined multiple point cloud serialization methods and bidirectional sequential modeling. We constructed a data-dependent local-global integrated state space model(SSM) segmentation model.
    \item Serialized Point Mamba possesses modeling capabilities equivalent to Transformer-based point cloud segmentation models while featuring linear complexity and computational demand.
    \item Serialized Point Mamba achieved  76.8 mIoU on the Scannet dataset and 70.3 mIoU on the S3DIS dataset. In instance segmentation on the Scannetv2 dataset, Serialized Point Mamba  achieved 40.0 mAP,outperforming comparable CNN and Transformer models. Serialized Point Mamba not only had the lowest latency but also maintained a reasonable memory footprint among all compared models. it is the SOTA among all point semantic segmentation models based on mamba.
\end{enumerate}

\section{Related Works}
\label{sec:RelatedWorks}



\subsection{Point Cloud Transformer}
Attention mechanisms aim to optimize a model's focus on different parts of an input sequence, allowing for more nuanced information processing. Compared to traditional deep learning networks, the Transformer\cite{vaswani2017attention} model, based on self-attention mechanisms, effectively models dependencies in long sequences. It has proven its prowess not only in NLP but also in two-dimensional visual domains.

In the realm of point cloud learning, in 2020, Guo et al. introduced the Point Cloud Transformer\cite{guo2021pct}, bringing in self-attention to enable the modeling of global context and relationships between points in point cloud data. Also in 2020, Zhao et al. proposed the Point Transformer\cite{zhao2021point}, differing from the Point Cloud Transformer by executing self-attention within local neighborhoods. In 2021, Yu X et al. introduced Point-BERT\cite{yu2022point}, performing unsupervised pretraining on standard Transformers trained on point clouds through masked reconstruction. Subsequent extensions of Transformers on point clouds demonstrated their tremendous potential across various downstream tasks in point cloud processing.

In 2022, Wu et al. proposed Point Transformer v2\cite{wu2022point}, utilizing vector Transformers and non-overlapping grid pooling to achieve state-of-the-art results on several point cloud understanding tasks. In 2023, Wang P S et al. introduced Octformer\cite{wang2023octformer}, organizing point cloud data using octrees and employing multi-head self-attention , achieving top-tier performance on the Scannet\cite{dai2017scannet} point cloud segmentation task. The same year, Yang Y Q proposed Swin3D\cite{yang2023swin3d}, extending the sliding window mechanism from Swin Transformer\cite{liu2021swin} to the point cloud domain, reducing the model's resource consumption.

In our work, we model point clouds using state-space models rather than attention mechanisms, inspired by these studies to perform multi-level modeling of point clouds. Our proposed Serialized Point Mamba (State-space Point Cloud Model) is capable of matching or surpassing current Transformer levels.

\subsection{State Space Model}
Originating from control theory\cite{kalman1960new}, state-space models (SSMs) that can now be integrated with deep learning for effective sequence modeling have emerged as a promising architecture for sequence modeling. These models can be interpreted as a combination of recurrent neural networks and convolutional neural networks.

In 2021, Gu et al. proposed Structured State-Space Models (S4)\cite{gu2021efficiently}, utilizing Hippo matrices for efficient compression and retention of past information, while introducing reparameterization techniques to reduce computational complexity, demonstrating significant potential in problems involving long-distance sequences. In 2022, Smith et al. introduced S5\cite{smith2022simplified} based on the S4 model, replacing SISO SSMs with MIMO and introducing parallel scanning to accelerate computation. The same year, Fu et al. proposed the H3 model\cite{fu2022hungry}, employing a new SSM layer using fast Fourier transforms to speed up state-space model training. In 2023, Peng et al. proposed RWKV\cite{pengrwkv}, combining Transformers with state memory using Time-mix and Channel-mix layers, achieving exceptionally high performance in handling sequential data.

Beyond natural language processing, various variants of SSMs\cite{orvieto2023resurrecting}\cite{smith2022simplified} have succeeded in domains involving continuous signal data such as audio\cite{goel2022s}\cite{saon2023diagonal}and vision\cite{nguyen2022s4nd}. 2D SSMs\cite{baron20232}, SGConvNeXt\cite{li2022makes}, and ConvSSM\cite{smith2024convolutional} combine SSMs with CNN or Transformer architectures to handle two-dimensional data. However, they perform poorly when modeling discrete and information-dense data like text.

In 2023, Gu et al. introduced Mamba\cite{gu2023mamba}, designing a data-dependent SSM layer and constructing a universal language model backbone, Mamba, which outperforms Transformers of various sizes on large-scale real-world datasets while scaling linearly in terms of computational load and memory demand with sequence length. Recent works have explored the potential of Mamba in the realms of images and videos. In 2024, Zhu proposed VisionMamba\cite{zhu2024vision}, adopting a bidirectional modeling approach to extend the Mamba model to the image domain for the first time. Building upon this, Ruan\cite{ruan2024vm} et al. and Yang\cite{yang2024vivim} et al. respectively proposed Mamba-based models for medical image segmentation and video segmentation. We demonstrated the potential of SSMs in point clouds, extending the Mamba model to the point cloud domain with improved outcomes.

In 2024,Liang\cite{liang2024pointmamba}  et al. and zhang\cite{zhang2024point}  et al. respectively proposed PointMamba.PointMamba employs a linear complexity algorithm, presenting global modeling capacity while significantly reducing computational costs. Specifically, their method leverages space-filling curves for effective point tokenization and adopts an extremely simple, non-hierarchical Mamba encoder as the backbone.Comprehensive evaluations demonstrate that PointMamba achieves superior performance across multiple datasets while significantly reducing GPU memory usage and FLOPs. This work underscores the potential of SSMs in 3D vision-related tasks and presents a simple yet effective Mamba-based baseline for future research.

\section{Method}
\label{sec:Method}
\subsection{Preliminaries}

\paragraph{State Space Models.}
The State Space Model (SSM) is defined by Eq.\ref{eq1}, projecting a one-dimensional input signal \(u(t)\) onto an N-dimensional latent state \(x(t)\), which is then projected back to a one-dimensional output signal.
\begin{equation}
\label{eq1}
\begin{split}
x'(t)&=Ax(t)+Bu(t)\\
y(t)&=Cx(t)+Du(t)
\end{split}
\end{equation}
Here, A, B, C, and D are parameters learned via gradient descent. Assuming \(D=0\), it can be considered a simple treatment with a residual connection. For a discrete sequence \((u_0, u_1, \ldots)\) instead of the continuous input \(u(t)\), Eq.\ref{eq1} must be discretized. Conceptually, the input \(u_k\) can be viewed as a sample from an underlying continuous signal \(u(t)\), where \(u_k = u(k\Delta)\), \(\Delta\) being the discretization step size. The discrete SSM is thus:
\begin{equation}
\label{eq2}
\begin{split}
x_k&=\bar{A}x_{k-1} + \bar Bu_k \\
y_k &= \bar{C}x_k + \bar{D}u_k
\end{split}
\end{equation}
Using bilinear or zero-order hold (ZOH) methods, the parameters A and B are transformed into approximations \(\bar{A}\) and \(\bar{B}\):
\begin{equation}
\begin{split}
\bar{A} &= \exp(\Delta A) \\
\bar{B} &= (\Delta A)^{-1}(\exp(\Delta A) - I) \cdot \Delta B
\end{split}
\end{equation}
To enhance computational efficiency, Eq.~\ref{eq2} can be accelerated using parallel methods, formulated in a global convolution manner:
\begin{equation}
\overline{\mathbf{K}} = (\mathbf{C}\overline{\mathbf{B}}, \mathbf{C}\overline{\mathbf{A}}\overline{\mathbf{B}}, \ldots, \mathbf{C}\overline{\mathbf{A}}^{M-1}\overline{\mathbf{B}}), \quad y = x \ast \overline{\mathbf{K}}.
\end{equation}
where \(M\) is the length of the input sequence \(x\), and \(\overline{\mathbf{K}}\) is the kernel for the global convolution.

\paragraph{Selective SSM.}
Although the SSM addresses linear complexity, the static state representation in SSMs can significantly impact model performance. The S6 algorithm, introduced in Mamba, transforms the linear time-invariant parameters in SSMs into input-dependent parameters. In S6, parameters \(\Delta\), B, and C are converted into input functions using learnable fully connected layers to project the number of channels to \(N\). A tensor dimension denoting the sequence length \(L\) is added, making the model data-dependent. Matrix A remains unchanged but is dynamically influenced through B and C.

\subsection{Serialized Point Cloud Mamba}

\paragraph{Serialization strategy.}
Unordered point cloud data must be organized into a one-dimensional sequence with some pattern or meaning for sequential modeling. This paper employs space-filling curves to reorganize the point cloud into a one-dimensional sequence. For sparse point clouds, grid sampling partitions the point cloud space. Points along the space-filling curve path in the grid's intervals have their projected positions recorded to generate a mapping list. The Z-order curve, Hilbert curve, and their variants "Trans-Hilbert" and "Trans-Z" are utilized. After serializing the point cloud, points adjacent in the new arrangement may also be spatially adjacent.
\begin{figure}
    \centering
    \includegraphics[scale=0.7]{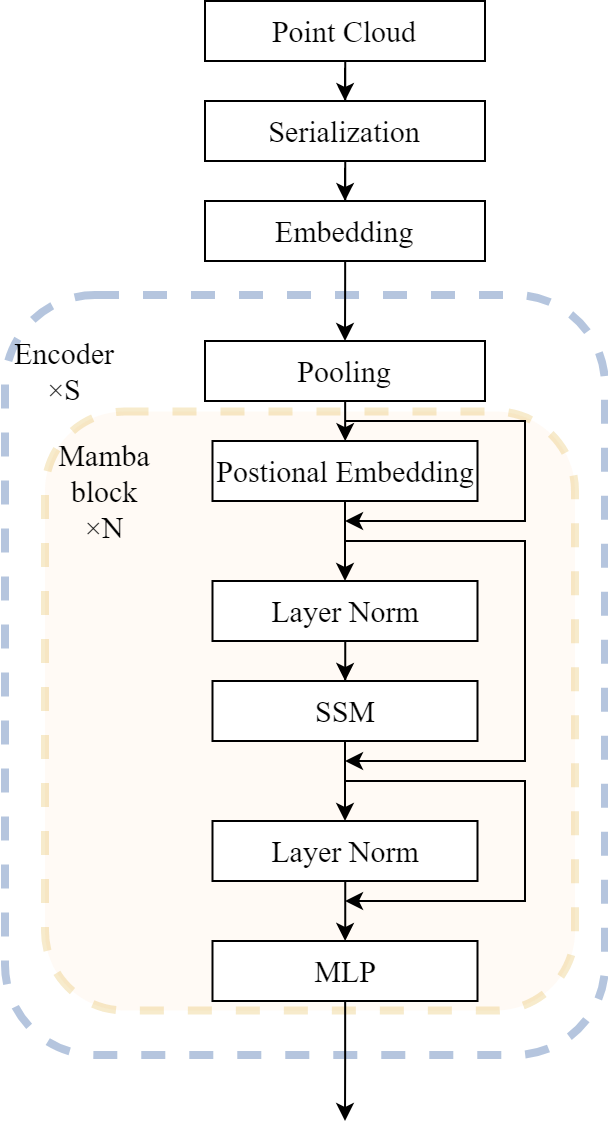}
    \caption{The architecture of Serialized Point Mamba encoder}
    \label{fig:encoder}
\end{figure}

\paragraph{Serialized Mamba.}
After point cloud serialization, the entire point cloud can theoretically be modeled. However, for large point cloud datasets like S3DIS \cite{armeni20163d}, where individual scene point clouds contain over 100,000 points, processing speeds are significantly impacted despite Mamba's parallel scanning algorithms. Inspired by Swin Transformer \cite{liu2021swin}, the point cloud sequence is divided into smaller subsequences of equal length, with Mamba applied to perform local modeling on each subsequence. For the remaining portion of the point cloud sequence insufficient to form a new subsequence, points from the nearest preceding subsequence are replicated, padding it to create a new subsequence of the same length. This division enables GPUs to process multiple subsequences concurrently, enhancing parallel efficiency. A multi-level scaling structure is utilized, gradually expanding the 'receptive field' of the subsequences until global modeling is achieved.

\begin{figure}
    \centering
    \includegraphics[scale=0.5]{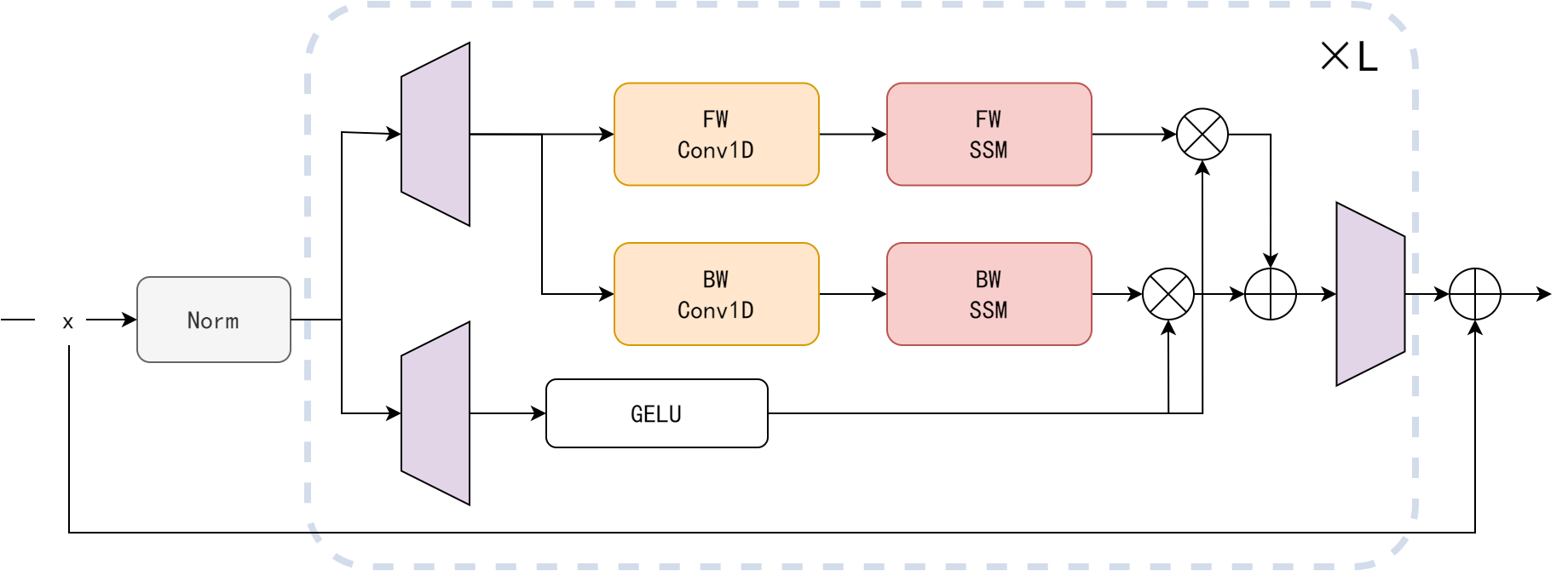}
    \caption{Bidirectional Mamba.}
    \label{fig:bimamba}
\end{figure}

\begin{figure}
    \centering
    \includegraphics[scale=0.8]{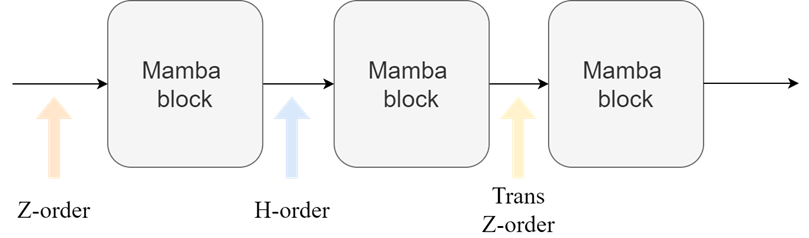}
    \caption{Multi-serialization methods utilization.}
    \label{fig:mulitorder}
\end{figure}

\paragraph{Subsequence interaction.}
Each subsequence is processed independently without interaction between them. To stabilize the model, bidirectional sequential modeling methods, such as bidirectional LSTM \cite{huang2015bidirectional}, can be referenced. Bidirectional modeling aids in capturing longer-range dependencies, enhancing prediction accuracy and the model's ability to grasp complex data patterns. However, it increases computational complexity. Inspired by VisionMamba \cite{zhu2024vision}, as illustrated in Figure~\ref{fig:bimamba}, after normalization, the input is projected using a linear layer and split into two branches of identical dimensions. One branch undergoes regular Mamba computation, involving one-dimensional convolution followed by SSM operations. The other branch reverses the sequence, then performs one-dimensional convolution and SSM operations. After calculations, the reversed branch is reverted to the normal sequence. Both branches are gated by a branch activation function (element-wise multiplication) before being concatenated along the channel direction and passed through a linear layer to project back to the input dimension, ensuring the input and output dimensions are the same, with the residual from before normalization added back in.
Another approach involves utilizing multiple serialization methods rather than relying on a single method to process the point cloud. As depicted in Figure~\ref{fig:mulitorder}, each Mamba block at every layer employs a different point cloud serialization method. Here, subsequences vary for each Mamba block, indirectly facilitating interaction among different subsequences.

\paragraph{Positional Embedding.}
Positional Encoding incorporates positional information, enabling the model to understand the ordinal relationships between elements in the sequence. Octformer \cite{wang2023octformer} adopted octree-based convolutional positional encoding for point clouds. To simplify, we utilize sparse submanifold convolutions \cite{spconv2022} to learn conditional positional encodings instead, computed as:
\begin{equation}
\label{eq:positonembedding}
x = x + norm(SubMConv3d(x))
\end{equation}
Here, \(x\) denotes the input point cloud, \(SubMConv3d\) represents the 3D sparse submanifold convolution, and \(norm\) signifies the normalization layer. Given multiple serialization methods, conditional positional encoding is applied before each Mamba block to enhance positional representations.

\paragraph{Grid Pooling.}
Pooling reduces spatial dimensions, decreasing the width and height of input feature maps while preserving critical feature information. Traditional sampling-based pooling methods combine sampling and querying. During sampling, farthest point or grid sampling selects points for the next encoding stage. Each sampled point undergoes a neighborhood query to aggregate information from adjacent points. Inspired by PTv2 \cite{wu2022point}, we adopt grid pooling, applying spatial pooling to the point cloud sequence, ensuring systematic and controlled reduction of spatial dimensions, maintaining consistent information density across pooled regions.

\paragraph{Architecture Details.}
A U-Net type architecture, composed of an encoder and decoder, is adopted, as illustrated in Figure~\ref{fig:encoder}. Following embedding, the serialized point cloud is fed into the encoder, consisting of five stages. Each layer initiates with a pooling operation, followed by conditional positional encoding and layer normalization. Selective state space model computations are performed, followed by another round of layer normalization and MLP calculations. Model details are presented in Table \ref{tab:model detail}.
\begin{table} [H]
 \caption{Model Detail}
  \centering
  \begin{tabular}{lll}
    \toprule
    Architectural     & Serialized Point Mamba-tiny     & Serialized Point Mamba \\
    \midrule
    Bidirectional Mamba  & True  & False     \\
    Mamba Layer     & (2, 2, 2, 6, 2) & (2, 2, 2, 6, 2)     \\
    Channel     & (192, 192, 192, 192, 192)       & (32, 64, 128, 256, 512)  \\
    Serialization &["z", "z-trans", "hilbert", "hilbert-trans"]&["z", "z-trans", "hilbert", "hilbert-trans"]\\
    SubSequence Length &(1024, 1024, 1024, 1024, 1024)&(1024, 1024, 1024, 1024, 1024)\\
    \bottomrule
  \end{tabular}
  \label{tab:model detail}
\end{table}

\section{Experiments}
\subsection{Experiments Settings}
We conducted benchmark testing for segmentation on the indoor datasets Scannet\cite{dai2017scannet} and S3DIS\cite{armeni20163d}, as well as the outdoor dataset nuScenes\cite{caesar2020nuscenes}. All experiments were conducted on an Nvidia 3090 Ti GPU. We trained the Serialized Point Mamba using the AdamW optimizer, with an initial learning rate of 5e-4, a learning rate warm-up for two epochs, employing the OneCycleLR learning rate strategy with a cosine cycle. The batch size was set to 12. The point cloud data used for training was augmented using common point cloud segmentation data augmentation strategies, incorporating random rotation, random flipping, and random scaling of the point cloud data. For datasets containing RGB data in the original point clouds, random jittering was applied to induce random offsets in the RGB features. Due to differences in indoor and outdoor environments resulting in varying point cloud sparsity, our experimental setup adhered to conventional practices\cite{pointcept2023}. Specifically, the sparse voxelization grid sampling size was set to 0.02 for indoor datasets, whereas for outdoor datasets, it was adjusted to 0.06. This differentiation in parameter settings acknowledges the inherent disparities in point density and ensures effective processing and analysis across diverse environments.
\subsection{Main Result}
\paragraph{Semantic segmentation.}
Table~\ref{tab:scannet seg} presents the semantic segmentation results of Serialized Point Mamba on the Scannet dataset, alongside comparisons with other representative methods. Among these, Serialized Point Mamba-tiny is a compact version of Serialized Point Mamba, utilizing bidirectional sequential modeling. This smaller parameter variant of the Serialized Point Mamba model is achieved by adjusting the number of Mamba blocks per level and modifying the channel counts. Conversely, Serialized Point Mamba employs unidirectional sequential modeling. Further insights into the comparative analysis of these modeling approaches can be found in Section~\ref{Ablation stydy}  through our ablation studies. Notably, Serialized Point Mamba delivers outstanding mIoU metrics on the Scannet validation set.
\begin{table}[H]
 \caption{Scannet semantic segmentation.}
  \centering
  \begin{tabular}{lc}
    \toprule
    Model     & Val(mIoU) \\
    \midrule
    PointNet++\cite{qi2017pointnet++} & 53.5\\
    PointNext\cite{qian2022pointnext} & 71.5\\
    PointConv\cite{wu2019pointconv} & 61.0\\
    SparseConvNet\cite{graham20183d} & 69.3\\
    MinkowskiNet\cite{choy20194d} & 72.2\\
    O-CNN\cite{wang2017cnn}  &74.0\\
    \midrule
    Point Transformer(PTv1)\cite{zhao2021point} & 70.6 \\
    Point Transformer v2\cite{wu2022point} &75.4\\
    OctFormer\cite{wang2023octformer}  & 74.5\\
    Swin3D\cite{yang2023swin3d} & 74.2\\
    \midrule
    Serialized Point Mamba-tiny&	75.94\\
    Serialized Point Mamba&	76.8\\
    \bottomrule
  \end{tabular}
  \label{tab:scannet seg}
\end{table}
Table~\ref{tab:s3dis seg} showcases the segmentation result of Serialized Point Mamba on the S3DIS indoor dataset, specifically Area5. To ensure fairness, all compared models were trained from scratch without utilizing additional training data or pretrained weights. Given that scenes in the S3DIS dataset are considerably larger than those in Scannet, this may pose challenges for sequence modeling approaches. Table~\ref{tab:nuScenes seg} exhibits the semantic segmentation outcomes of Serialized Point Mamba on the nuScenes dataset, along with comparative analyses. Among the models evaluated, Cylinder3D and SphereFormer, which are projection-based methods, also exemplify state-of-the-art performance on outdoor datasets. Notably, Serialized Point Mamba outperforms all compared models, achieving the best results across the board.
\begin{table}[H]
 \caption{S3DIS semantic segmentation.}
  \centering
  \begin{tabular}{lc}
    \toprule
    Model     & Val(mIoU) \\
    \midrule
MinkUnet\cite{choy20194d}&65.4\\
PointNeXt\cite{qian2022pointnext}&70.5\\
PTv1\cite{zhao2021point}&70.4\\
PTv2\cite{wu2022point}&71.6\\
Serialized Point Mamba &70.6\\
    \bottomrule
  \end{tabular}
  \label{tab:s3dis seg}
\end{table}
\begin{table}[H]
 \caption{nuScenes semantic segmentation.}
  \centering
  \begin{tabular}{lc}
    \toprule
    Model     & Val(mIoU) \\
    \midrule
MinkUNet\cite{choy20194d}&73.3\\
SPVNAS\cite{tang2020searching}&77.4\\
Cylender3D\cite{zhu2021cylindrical}&76.1\\
SphereFormer\cite{lai2023spherical}&78.4\\
PTv2\cite{wu2022point}&80.2\\
Serialized Point Mamba	&80.6\\
    \bottomrule
  \end{tabular}
  \label{tab:nuScenes seg}
\end{table}
\paragraph{Instance Segmentation.}
Furthermore, to explore the scalability of our model, instance segmentation experiments were conducted on the Scannetv2 dataset, employing PointGroup\cite{jiang2020pointgroup} as the segmentation head. The instance segmentation model was fine-tuned using pre-trained weights from the semantic segmentation model trained on the Scannetv2 dataset. Table ~\ref{tab:scannet insseg} presents the results of different backbone networks for instance segmentation using PointGroup, where the instance segmentation performance of Serialized Point Mamba surpasses comparative point cloud CNN models and point cloud Transformer models.

\begin{table}[H]
 \caption{Scannet instance segmentation.}
  \centering
  \begin{tabular}{lccc}
    \toprule
Backbone &	mAP&	mAP25&	mAP50\\
    \midrule

MinkUnet\cite{choy20194d} &36.0	&72.8	&56.9\\
PTv2\cite{wu2022point} &38.3&	76.3	&60.0\\
Serialized Point Mamba	&40.0&	76.4	&61.4\\
    \bottomrule
  \end{tabular}
  \label{tab:scannet insseg}
\end{table}
\paragraph{Model Efficiency.}
We evaluate the operational efficiency of our model by analyzing average latency and memory consumption on real-world datasets. Efficiency metrics are measured on a single NVIDIA GeForce RTX 3090 Ti GPU, excluding the first iteration to ensure steady-state measurements, using the same optimizer and optimization strategy across models to maintain consistency. Results are averaged over one epoch to provide a reliable assessment. As shown in Table~\ref{tab: model effi}, we compare Serialized Point Mamba against a multitude of models, evaluating efficiency using the Scannetv2 dataset. Our findings indicate that Serialized Point Mamba not only boasts the lowest latency among all compared models but also maintains a reasonable memory footprint, all while outperforming competing models in terms of segmentation accuracy. Due to the increased memory consumption and inference time attributed to bidirectional sequence learning in Serialized Point Mamba-tiny, we opt for the Serialized Point Mamba model in subsequent experiments for a more balanced performance profile.
\begin{table}[H]
\caption{Model Efficiency}
\centering
\begin{tabular}{lccccc}
\hline
\multirow{2}{*}{Models} & \multirow{2}{*}{Params/m} & \multicolumn{2}{c}{Training} & \multicolumn{2}{c}{Inference} \\ \cline{3-6} 
                        &                           & Latency/ms     & Memory/G    & Latency/ms     & Memory/G    \\ \hline
MinkUnet{[}11{]}        & 37.9                      & 286            & 5.2         & 112            & 4.6          \\
OctFormer{[}25{]}       & 44                        & 295            & 12.9        & 114            & 12.5         \\
Swin3D{[}27{]}          & 71.1                      & 780            & 13.6        & 749            & 8.7          \\
PTV2{[}24{]}            & 12.8                      & 328            & 13.4        & 247            & 18.1         \\
Serialized Point Mamba-tiny               & 16                        & 196            & 10.8        & 169            & 9.6          \\
Serialized Point Mamba               & 50                        & 206            & 5.6         & 99             & 4.4          \\ \hline
\end{tabular}
\label{tab: model effi}
\end{table}
\subsection{Ablation Study}
\label{Ablation stydy}
\paragraph{Bidirecitonal Mamba.}
As shown in the contrast experiment in Table~\ref{tab:Bidirecitonal}, which compares bi-directional encoding(mIoU) on the Scannetv2 dataset. In models with fewer parameters, bi-directional sequential modeling can significantly improve segmentation accuracy by approximately five percentage points, demonstrating that bi-directional sequential modeling can markedly enhance the model's learning ability with a smaller number of parameters. However, for models with larger parameter counts, the effect of bi-directional sequential modeling is not as pronounced. Different serialization modes result in different point cloud sequences. Here, we speculate that in models with more parameters, due to the mixed use of various serialization modes and the greater number of channels for learning different point cloud serialization patterns, the model's generalization ability is enhanced even more. To some extent, this indirectly achieves the effect of bi-directional sequential modeling.
\begin{table}[H]
 \caption{Bidirectional Encoding.}
  \centering
  \begin{tabular}{lcc}
    \toprule
    Bidirectional     & True & False \\
    \midrule
Serialized Point Mamba-tiny	&75.94&	72.3\\
Serialized Point Mamba&	76.67	&76.8\\
    \bottomrule
  \end{tabular}
  \label{tab:Bidirecitonal}
\end{table}
\paragraph{Serialization.}
The Serialized Point Mamba-base model with a larger number of parameters was used to compare serialization patterns on the Scannetv2 dataset, as shown in Table~\ref{tab:seri}. Utilizing multiple serialization modes significantly enhances the model's learning capability, confirming the effectiveness of multi-serialization modes. Moreover, adopting a randomized serialization mode approach further improves the model's generalization ability.
\begin{table}[H]
\centering
\caption{Serialization Pattern.}
\begin{tabular}{lcccc}
\hline
Serialization Pattern         & \multirow{2}{*}{H} & \multirow{2}{*}{H+Z} & \multirow{2}{*}{H+transH+z+transZ} & \multirow{2}{*}{+Random Order} \\ \cline{1-1}
Model            &                    &                      &                           &                        \\ \hline
Serialized Point Mamba/mIoU(\%) & 74.64              & 75.1                 & 76.37                     & 76.8   \\ \hline              
\end{tabular}
\label{tab:seri}
\end{table}
\paragraph{Positional Embedding.}
The ablation study for positional encoding is presented in Table.~\ref{tab:PE}, conducted on the Scannetv2 dataset. CPE stands for Conditional Positional Encoding, while Enhanced CPE refers to the application of new conditional positional encoding before each serialization modeling operator to strengthen the positional representation.
Even without standalone positional encoding, an mIoU of 69.75\% can be achieved. As discussed in section 3.1.4, this is because the non-downsampling sparse submanifold convolution used in the embedding encoding phase indirectly serves the function of conditional positional encoding.
Using enhanced conditional positional encoding improves the segmentation metric by 0.6 percentage points compared to employing CPE only once at each downsampling stage. This demonstrates the effectiveness of utilizing enhanced conditional positional encoding.
\begin{table}[H]
 \caption{Positional Encoding.}
  \centering
  \begin{tabular}{lc}
    \toprule
Positinal Encoding &	mIoU/\% \\
    \midrule
No PE	&69.75 \\
CPE	&76.21 \\
Enhanced CPE & 76.8 \\
    \bottomrule
  \end{tabular}
  \label{tab:PE}
\end{table}
\paragraph{Norm Layer.}
The experiments on the position of normalization layers within the Block, as shown in Table~\ref{tab:norm-pos} conducted on the Scannetv2 dataset, compare the segmentation metrics of models using pre-normalization and post-normalization structures. The results demonstrate that the pre-normalization model performs significantly better than the post-normalization structure. This suggests that during the training process, it is crucial to first normalize the point cloud data to adjust its distribution before feeding it into the respective operators.
\begin{table}[H]
 \caption{Norm Layer Position.}
  \centering
  \begin{tabular}{lc}
    \toprule
    Block Structure     & mIoU/\% \\
    \midrule
Pre-Norm	&76.8\\
Post-Norm&	73.35	\\
    \bottomrule
  \end{tabular}
  \label{tab:norm-pos}
\end{table}
\paragraph{Subsequence Length.}

In our proposed model, all Mamba blocks have the same subsequence size. In different point cloud sequences, if the remaining sequence length is insufficient to form a new subsequence, the previous subsequence is used for padding. Subsequence division operations are performed only when the total size of the point cloud exceeds the sequence length. As shown in Table~\ref{tab:se-le}, it presents the segmentation results of the Serialized Point Mamba model on the Scannetv2 dataset with varying sequence lengths. It is evident that selecting an appropriate subsequence length significantly enhances the model's learning of point cloud sequences.
\begin{table}[H]
 \caption{Subsequence Length.}
  \centering
  \begin{tabular}{lc}
    \toprule
    Subsequence Length    & mIoU/\% \\
    \midrule
512&	75.83 \\
1024&	76.8\\
2048&	76.23\\
    \bottomrule
  \end{tabular}
  \label{tab:se-le}
\end{table}

\section{Conclusion}
We have introduced the Serialized Point Mamba, a novel point cloud segmentation model based on Mamba. By applying the Mamba model to point cloud data and serializing it using space-filling curves, we have enhanced the model's ability to capture point cloud structures and generalize, thus improving segmentation outcomes. 

A staged point cloud sequence learning method was proposed, enabling effective handling of large-scale point cloud data by utilizing local sequence modeling. This approach reduced the model's parameter count and computational load. Additionally, grid pooling was employed to address issues of overlapping downsampling ranges, spatially aligning pooled points to preserve the spatial structural information of the point cloud. This method avoided complex sampling and neighbor query steps, thus increasing the efficiency of pooling operations.

We adopted Conditional Positional Encoding to dynamically encode positional information based on the input point cloud sequence, using multiple such encodings to accommodate different space-filling curve serialization methods. Through the dynamic compression of point cloud sequences with the state-space model Mamba, Serialized Point Mamba effectively reduced memory usage and computation, enhancing the efficiency and accuracy of point cloud semantic segmentation.

Experimental results demonstrate that Serialized Point Mamba outperforms existing technologies on public datasets such as ScanNetv2, S3DIS, and nuScenes. These results underscore the model's capability to deliver state-of-the-art performance across diverse point cloud segmentation tasks.


\bibliographystyle{unsrt}  
\bibliography{references}  

\end{document}